\documentclass{article}

\usepackage[preprint]{neurips_2026}

\usepackage[utf8]{inputenc} 
\usepackage[T1]{fontenc}    
\usepackage{hyperref}       
\usepackage{url}            
\usepackage{booktabs}       
\usepackage{amsfonts}       
\usepackage{nicefrac}       
\usepackage{microtype}      
\usepackage{xcolor}         
\usepackage[dvipsnames]{xcolor}
\usepackage{amsmath}

\usepackage{multirow}
\usepackage{graphicx}
\usepackage{subcaption}
\usepackage{wrapfig}

\usepackage{listings}
\usepackage{subcaption}
\usepackage{xcolor,colortbl}
\usepackage{bbm}
\usepackage{pifont}
\usepackage{caption}

\usepackage{amssymb}
\usepackage{graphicx}
\usepackage{calc}
\usepackage[most]{tcolorbox}

\title{From Tokens to Regions: CUDA-Sensitive Instruction Tuning for GPU Kernel Generation}

\newcommand*{\affaddr}[1]{#1} 
\newcommand*{\affmark}[1][*]{\textsuperscript{#1}}

\author{Wentao Chen\affmark[1]\quad Jiace Zhu\affmark[1]\quad Xing Zhe Chai\affmark[1]\quad Zeng Qu\affmark[2]\quad Qiaoling Xiao\affmark[2]\\\textbf{Liucheng Duan\affmark[2]}\quad \textbf{An Zou\affmark[1]\thanks{Corresponding Author.}}\\\affaddr{\affmark[1]Shanghai Jiao Tong University}\quad\affaddr{\affmark[2]Biren Technology}\\\texttt{\{wentaochen, zhujiace, xingzhechai, an.zou\}@sjtu.edu.cn}\\\texttt{\{zqu, qlxiao, aduan\}@birentech.com}\\
}

\begin{document}

\maketitle

\begin{abstract}
High-performance CUDA kernels are essential for scalable AI systems, while Large Language Models (LLMs) still struggle to generate correct kernels due to strict and implicit execution constraints. Existing LLM-based approaches either rely on costly agentic or reinforcement-learning (RL) pipelines, or adopt supervised fine-tuning (SFT) objectives that fail to explicitly model CUDA sensitivity, namely code tokens or regions tightly coupled with execution constraints. In this work, we investigate CUDA sensitivity from the perspective of token confidence patterns, showing that CUDA sensitivity appears at both token and region levels, where most CUDA-sensitive tokens are predicted with high confidence, while a smaller low-confidence subset forms regions corresponding to execution-critical structures. These findings suggest that effective CUDA kernel generation should both leverage high-confidence CUDA-sensitive tokens and preserve low-confidence CUDA-sensitive regions. Building on these insights, we propose \textbf{\underline{CU}DA-\underline{Se}nsitive Instruction \underline{T}uning (CuSeT)}, a low-cost post-training method within a simple SFT framework. CuSeT follows the principle of ``from tokens to regions'' by combining \emph{adaptive token-level masking} with \emph{region-aware sample reweighting}. Experiments show that CuSeT consistently improves functional correctness across multiple model families and scales, outperforming standard SFT and advanced SFT variants, while achieving competitive performance against frontier CUDA kernel generation models with substantially lower inference cost.
\end{abstract}

\section{Introduction}
\label{sec:intro}
High-performance GPU kernels are a cornerstone of scalable AI systems \citep{dao2022flashattention,ye2025flashinfer,miao2025towards}, yet developing such kernels remains challenging due to strict hardware-level execution constraints. While Large Language Models (LLMs) show promise in automating GPU kernel generation \citep{ouyang2025kernelbench}, existing approaches struggle to produce correct and executable CUDA code. These approaches include agentic frameworks \citep{chen2025cuda,wei2025astra,dong2025stark,saba2026cutegen}, post-training methods such as supervised fine-tuning (SFT) \citep{kernelllm2025,lv2025hpctranscompile,kong2025concur}, reinforcement learning (RL) \citep{li2025cuda,baronio2025kevin,liu2026dr}, as well as hybrid training pipelines that combine multiple stages \citep{woo2025tritonrl,cao2026ascendkernelgen,li2025autotriton}. As summarized in Tab. \ref{tab:compare}, existing paradigms generally lack CUDA sensitivity awareness, rely on execution feedback rather than CUDA-sensitive structural signals, or provide supervision only at token-level or whole-code-level granularity, often incurring additional inference or training cost.

\begin{table}[!ht]
\centering
\setlength{\tabcolsep}{5pt}
\begin{footnotesize}
\centering
\caption{Comparison of LLM-based GPU kernel generation methods.}
\begin{tabular}{ccccc}
\toprule
\multicolumn{2}{c}{\textbf{Paradigm}}                     & \begin{tabular}[c]{@{}c@{}}\textbf{CUDA Sensitivity}\\\textbf{Awareness}\end{tabular} & \begin{tabular}[c]{@{}c@{}}\textbf{CUDA-Sensitive}\\\textbf{Structural Signal}\end{tabular} & \begin{tabular}[c]{@{}c@{}}\textbf{Supervision}\\\textbf{Granularity}\end{tabular} \\ \midrule
\multicolumn{2}{c}{Agentic \citep{chen2025cuda,wei2025astra,dong2025stark,saba2026cutegen}}                      & \textcolor{Red}{No}                         & \textcolor{Orange}{Execution Feedback}              & \textcolor{Red}{Whole-code level}               \\ \midrule
\multirow{5}{*}{Post-training} & Standard SFT \citep{kernelllm2025,lv2025hpctranscompile,kong2025concur}    & \textcolor{Red}{No}                         & \textcolor{Red}{No}                          & \textcolor{Red}{Uniform token-level}                \\
                               & Improved SFT \citep{wu2025generalization,diao2026entropy,zhu2025anchored,liu2026profit}   & \textcolor{Red}{No}                         & \textcolor{Red}{No}                          & \textcolor{Red}{Token-level}                \\
                               & RL \citep{li2025cuda,baronio2025kevin,liu2026dr}             & \textcolor{Red}{No}                         & \textcolor{Orange}{Execution Feedback}              & \textcolor{Red}{Whole-code level}               \\
                               & Hybrid \citep{woo2025tritonrl,cao2026ascendkernelgen,li2025autotriton} & \textcolor{Red}{No}                         & \textcolor{Orange}{Execution Feedback}              & \textcolor{Red}{Whole-code level}               \\ \cmidrule{2-5}
                               & \textbf{CuSeT (Ours)}     & \textbf{\textcolor{Green}{Yes}}                        & \textbf{\textcolor{Green}{Region Confidence}}                & \textbf{\textcolor{Green}{Token-to-region}}                \\ \bottomrule
\end{tabular}
\label{tab:compare}
\end{footnotesize}
\end{table}

The fundamental reason behind this limitation is the mismatch between existing methods and CUDA programming. Unlike general-purpose code generation \citep{wei2024magicoder,roziere2023code}, CUDA kernel generation is a system-constrained conditional generation problem. A generated kernel must satisfy not only syntax and semantics, but also implicit execution constraints \citep{cook2012cuda,kirk2016programming}, including memory access patterns, thread indexing, synchronization, and launch configuration. We define \emph{CUDA sensitivity} to describe code tokens or regions whose correctness is tightly coupled with these execution constraints. Such CUDA-sensitive tokens are concentrated in execution-critical locations and regions, where small deviations may lead to compilation failures or incorrect execution, while many other tokens contribute far less to functional correctness.

This token-level and region-level sensitivity poses a fundamental challenge to existing CUDA kernel generation paradigms. Agentic methods refine kernels through iterative feedback, but they often require long optimization loops and substantial inference cost. RL-based methods optimize execution-level objectives, but their rewards are typically sparse and available only after execution. Hybrid pipelines combine multiple signals but introduce additional training and system complexity. Standard SFT remains an attractive foundation due to its simplicity, stability, and computational efficiency. However, it applies uniform supervision across all target tokens \citep{ouyang2022training} and does not explicitly account for token-level and region-level CUDA sensitivity, which limits its effectiveness in supervising execution-critical code. Existing SFT improvements partially address token importance by reweighting token losses using predicted probabilities, entropy, or other uncertainty signals \citep{wu2025generalization,diao2026entropy,zhu2025anchored}, but they remain insufficient for CUDA kernel generation as they are usually CUDA-agnostic and overlook regional structure. Consequently, an effective SFT-based approach for CUDA kernel generation should address CUDA sensitivity at two complementary granularities: strengthening supervision for execution-critical tokens while emphasizing samples containing low-confidence CUDA-sensitive regions.

To better understand how CUDA sensitivity manifests in model predictions, we analyze token confidence patterns in CUDA code, revealing that CUDA sensitivity occurs at both token and region levels: most CUDA-sensitive tokens are predicted with high confidence, while a smaller low-confidence subset forms regions corresponding to execution-critical structures. This finding motivates a supervision strategy from tokens to regions for CUDA kernel generation: leveraging high-confidence CUDA-sensitive tokens while preserving low-confidence CUDA-sensitive regions. 

Building on these insights, we propose \textbf{CUDA-Sensitive Instruction Tuning (CuSeT)}, a low-cost post-training method within a simple SFT framework that explicitly strengthens supervision over CUDA-sensitive tokens and regions. CuSeT implements the principle of ``from tokens to regions'' via two complementary designs: (1) \emph{adaptive token-level masking}, which filters noisy low-confidence CUDA-neutral tokens while preserving CUDA-sensitive tokens, and (2) \emph{region-aware sample reweighting}, which emphasizes samples containing low-confidence CUDA-sensitive regions. In this way, token-level masking determines which tokens contribute to supervision, while region-aware sample reweighting identifies difficult regions that require greater optimization emphasis. By explicitly modeling CUDA sensitivity at both token and region levels, CuSeT improves functional correctness while maintaining computational efficiency, without relying on costly RL or complex multi-stage pipelines. Our contributions are summarized as follows:

\begin{itemize}
    \item We introduce \textbf{CUDA sensitivity} and analyze its token-level and region-level patterns, showing that most CUDA-sensitive tokens are high-confidence, while a small subset of low-confidence tokens forms regions critical for execution.
    \item We propose \textbf{CUDA-Sensitive Instruction Tuning (CuSeT)}, a low-cost SFT-based post-training method that combines \textbf{adaptive token-level masking} and \textbf{region-aware sample reweighting} to strengthen supervision over CUDA-sensitive tokens and regions.
    \item Extensive experiments demonstrate that CuSeT significantly improves functional correctness across multiple model families and scales, outperforming standard SFT and improved SFT methods, while achieving competitive performance against frontier CUDA kernel generation models with substantially lower inference cost.
\end{itemize}

\section{Related Work}
\subsection{LLMs for GPU Kernel Generation}
Recently, LLMs have demonstrated remarkable general capabilities in software development tasks \citep{guo2024deepseek,roziere2023code}, but applying LLMs to GPU programming remains extremely challenging. Ouyang \emph{et al.} introduced KernelBench \citep{ouyang2025kernelbench}, showing that current LLMs struggle to generate both correct and performant GPU code, highlighting the gap between general code generation and domain-specific high-performance programming. Following works have investigated a variety of approaches to automate GPU kernel generation and optimization using LLMs. These mainly include leveraging agentic systems with self-refinement and iterative optimization \citep{chen2025cuda,lange2025ai,dong2025stark,lei2025pragma}. While effective in certain settings, these approaches incur substantial computational costs, requiring long iterative cycles, or complex multi-stage pipelines. Instead, more works have shifted towards training more powerful models capable of handling GPU kernel programming. These include supervised fine-tuning (SFT) methods \citep{kernelllm2025,kong2025concur,ke2025qimeng}, where pre-trained LLMs are adapted to GPU kernel generation tasks by fine-tuning on a domain-specific dataset. Additionally, reinforcement learning (RL)-based methods have also been explored \citep{baronio2025kevin,li2025cuda,liu2026dr}, which focus on optimizing the kernel generation process by rewarding the model for generating correct and efficient kernels. Some works \citep{li2025autotriton,woo2025tritonrl,cao2026ascendkernelgen} have combined SFT with RL to leverage the advantages of both techniques. However, despite these advancements, these methods often overlook the unique characteristics of CUDA code, including implicit execution constraints, as well as token-level and region-level CUDA sensitivity, limiting their ability to generate functionally correct and executable GPU kernels.

\subsection{Post-training LLMs}
Post-training methods, including SFT and RL, are widely applied to align pre-trained LLMs with specific tasks. Standard SFT minimizes token-level negative log-likelihood \citep{ouyang2022training}, assuming uniform token importance, while RL updates the model based on its generated outputs using task-specific reward signals \citep{shao2024deepseekmath,schulman2017proximal}. However, in CUDA code, execution-critical tokens are unevenly distributed, and uniform supervision fails to capture the contributions of CUDA-sensitive tokens and regions. Token-level improvements using prediction probabilities \cite{wu2025generalization,li2025beyond} or entropy \cite{diao2026entropy,wang2026gradients} provide uncertainty-aware supervision but remain token-centric and CUDA-agnostic. Similarly, RL-based approaches \citep{baronio2025kevin,li2025autotriton,li2025cuda} also face complementary challenges since sparse rewards, often available only at the final execution result, fail to provide fine-grained feedback on critical CUDA-sensitive regions. In this work, we analyze CUDA sensitivity from tokens to regions and propose \textbf{CUDA-Sensitive Instruction Tuning (CuSeT)}, a low-cost SFT-based post-training method that explicitly strengthens supervision over execution-critical CUDA-sensitive tokens and regions, improving functional correctness in CUDA kernel generation.

\section{Motivation}
\label{sec:motivation}
To investigate how CUDA sensitivity is reflected in model predictions, we select representative CUDA code samples from NVIDIA TensorRT \citep{nvidiatensorrt} and label tokens as \emph{CUDA-sensitive} or \emph{CUDA-neutral} based on their dependence on execution constraints, such as memory access patterns or thread indexing, as shown in Fig. \ref{fig:code_sample}. We then compute each token's predicted probability using Qwen3-8B \citep{yang2025qwen3}, referred to as its \emph{token confidence}. Our analysis reveals two key observations.

\begin{figure}[!ht]
    \centering
    \includegraphics[width=0.5\linewidth]{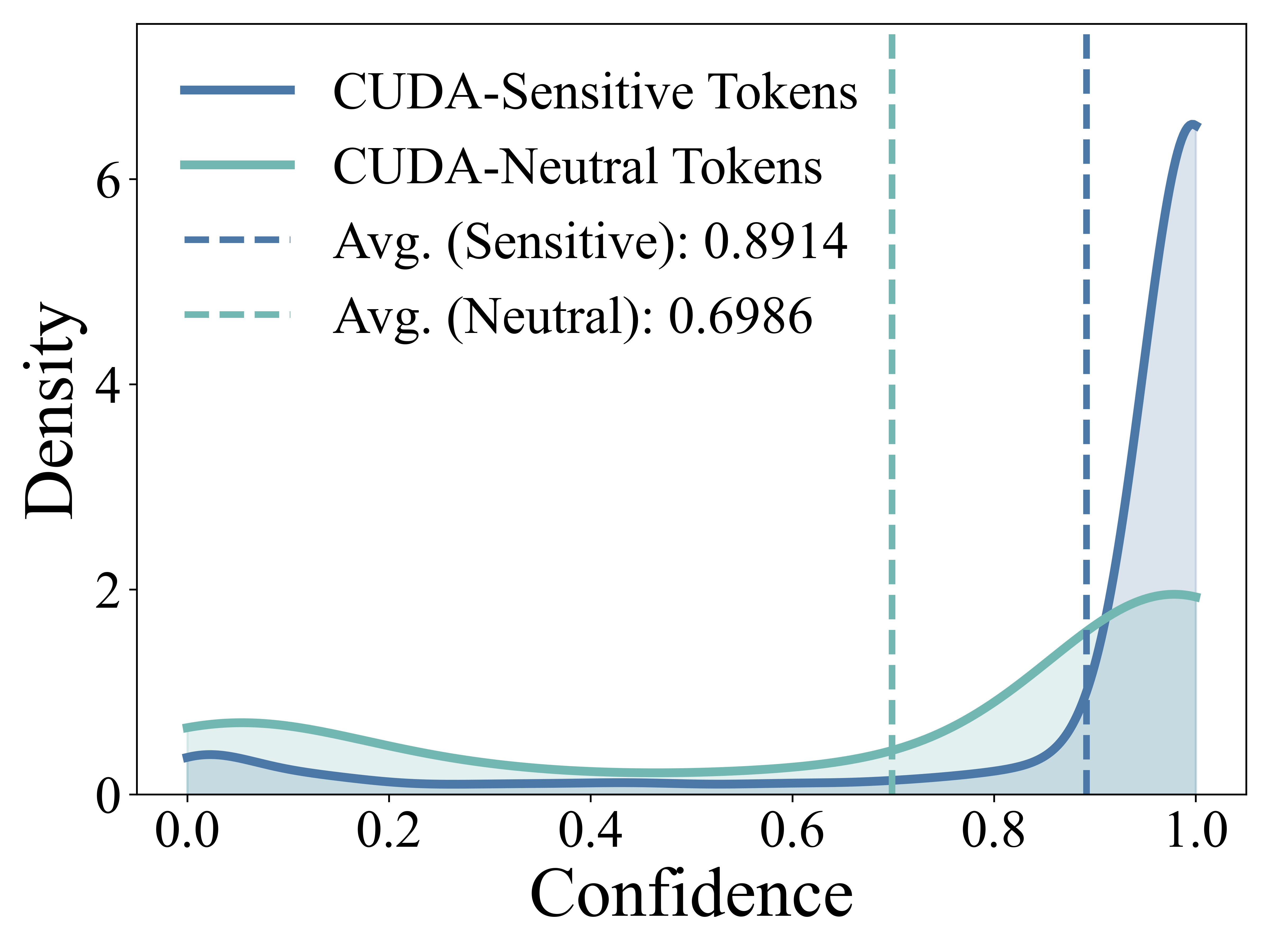}
    \caption{Token confidence density distributions of CUDA-sensitive and CUDA-neutral tokens in CUDA code, computed as the model's next-token prediction probabilities. CUDA-sensitive tokens encode execution-critical logic and concentrate in high-confidence regions, while CUDA-neutral tokens are more dispersed with higher density in low-confidence regions. The two distributions are statistically distinct under a two-sample KS test.}
    \label{fig:p}
\end{figure}

At the token level, CUDA-sensitive and CUDA-neutral tokens exhibit clearly different confidence distributions. As shown in Fig. \ref{fig:p}, most CUDA-sensitive tokens are concentrated in high-confidence regions, suggesting that token confidence can serve as a reliability signal for supervising execution-critical code. In contrast, CUDA-neutral tokens are more dispersed and have higher density in low-confidence regions, indicating less reliable supervision signals. A two-sample Kolmogorov-Smirnov (KS) test \citep{massey1951kolmogorov} confirms that the confidence distributions are statistically distinct\footnote{The null hypothesis assumes that CUDA-sensitive and CUDA-neutral tokens are drawn from the same confidence distribution. The test yields $p=4.6\times10^{-39}$, leading to a significant rejection of the null hypothesis.}. This observation suggests that token confidence can serve as a useful reliability signal for token-level supervision.

\begin{figure}[!ht]
    \centering
    \begin{subfigure}[!ht]{0.48\linewidth}
        \centering
        \includegraphics[width=\linewidth]{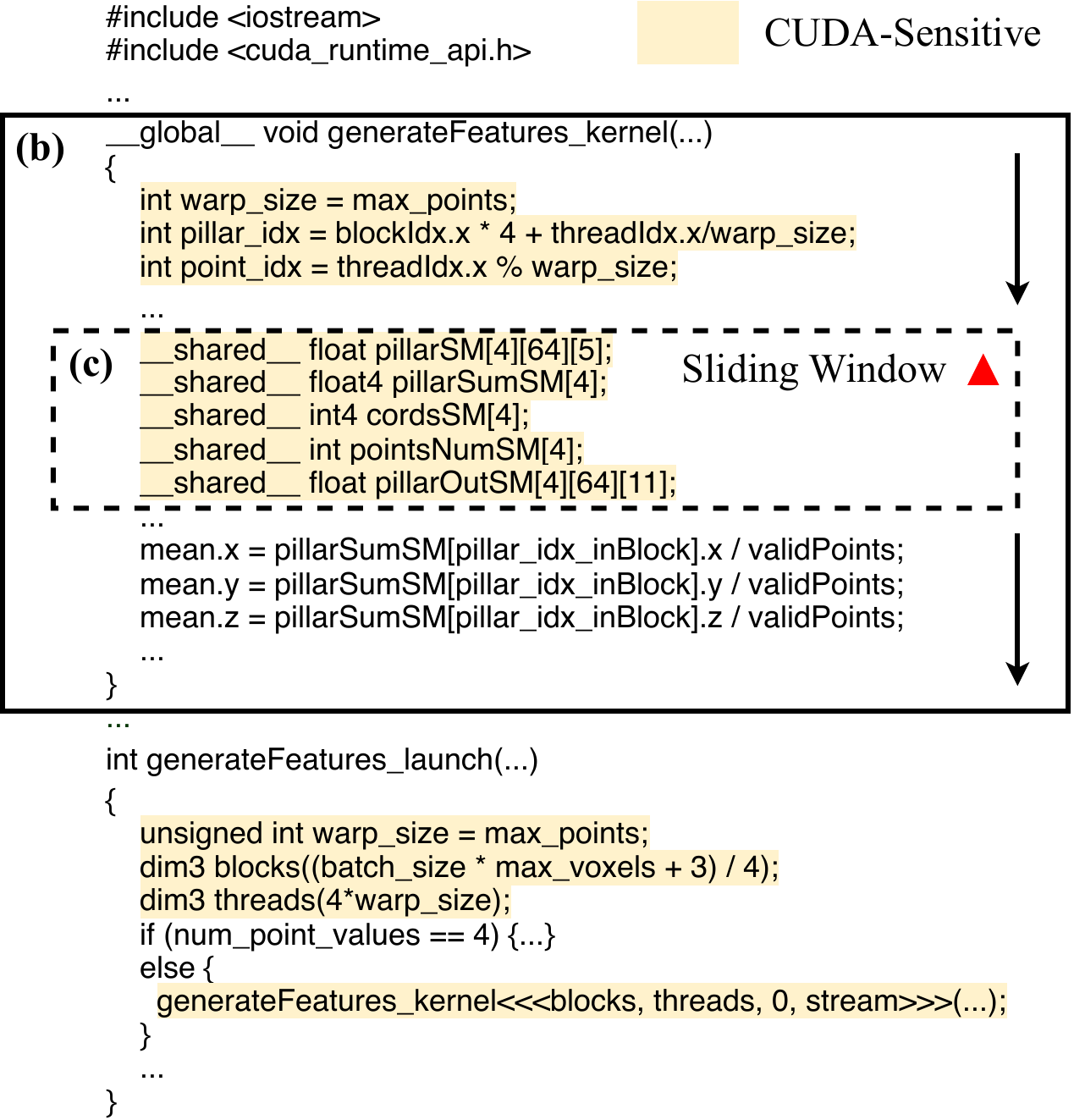}
        \caption{}
        \label{fig:code_sample}
    \end{subfigure}\hfill
    \begin{subfigure}[!ht]{0.5\linewidth}
        \centering
        \includegraphics[width=\linewidth]{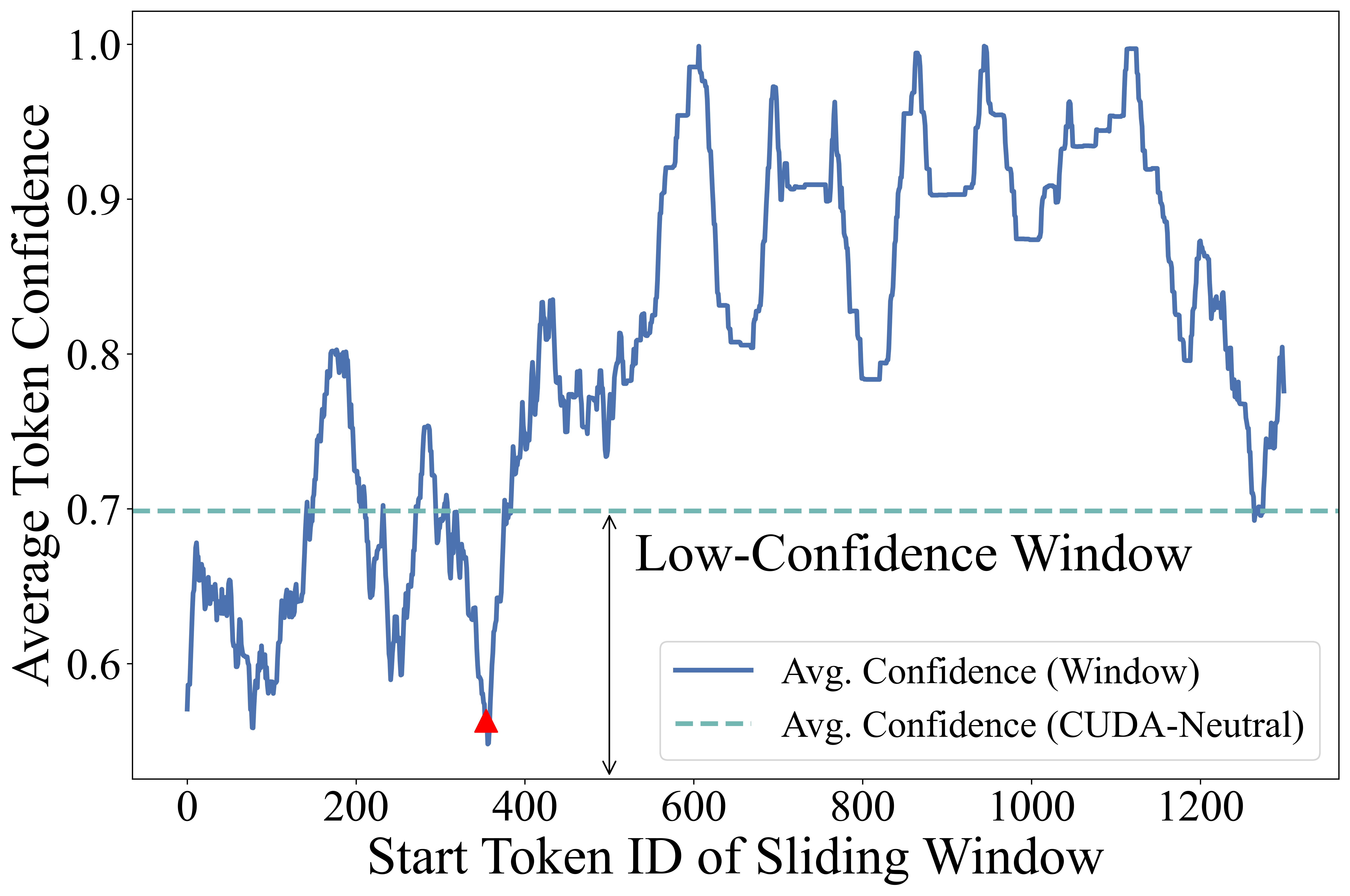}
        \caption{}
        \label{fig:confidence_curve}
    \end{subfigure}\hfill
    \begin{subfigure}[!ht]{\linewidth}
        \includegraphics[width=\linewidth]{fig/token_probabilities_zoom_in.png}
        \caption{}
        \label{fig:zoom_in_seq_p}
    \end{subfigure}
    \caption{(a) A sample CUDA code with CUDA-sensitive tokens highlighted; (b) Average token confidence over sliding windows within the solid-box region in (a), using a window size of 64 tokens and a stride of 1 token. Low-confidence windows are defined as sliding windows whose average token confidence is below the average confidence of CUDA-neutral tokens in Fig.~\ref{fig:p}; (c) Zoom-in view of a representative low-confidence window ($\textcolor{red}{\blacktriangle}$), showing individual token confidences. Across all samples, \textbf{74.47\%} contain at least one low-confidence window, and \textbf{91.18\%} of these low-confidence windows correspond to CUDA-sensitive code structures, indicating that a low-confidence subset of CUDA-sensitive tokens forms regions associated with execution-critical structures.}
    \label{fig:midexp}
\end{figure}

At the region level, a small subset of low-confidence CUDA-sensitive tokens forms regions corresponding to execution-critical structures, such as shared memory allocation or launch configuration. Fig.~\ref{fig:midexp} illustrates one representative example. Unlike low-confidence tokens in general code generation tasks, which often reflect stylistic variation or interchangeable expressions with limited semantic impact, these regions indicate structured uncertainty about execution constraints. Properly supervising these regions is critical for learning difficult CUDA patterns.

Together, these observations indicate that supervision in CUDA kernel generation should go beyond token-level confidence: high-confidence CUDA-sensitive tokens provide reliable guidance, whereas a low-confidence subset should be preserved at the region level because it corresponds to execution-critical structures. This motivates a supervision strategy from tokens to regions, leveraging reliable high-confidence tokens while preserving low-confidence CUDA-sensitive regions. Such a strategy serves as the basis for our method, \textbf{CuSeT}, which explicitly incorporates CUDA sensitivity at both token and region levels.

\section{CUDA-Sensitive Instruction Tuning}
\subsection{Overall}
\begin{figure}[t]
    \centering
    \includegraphics[width=0.95\linewidth]{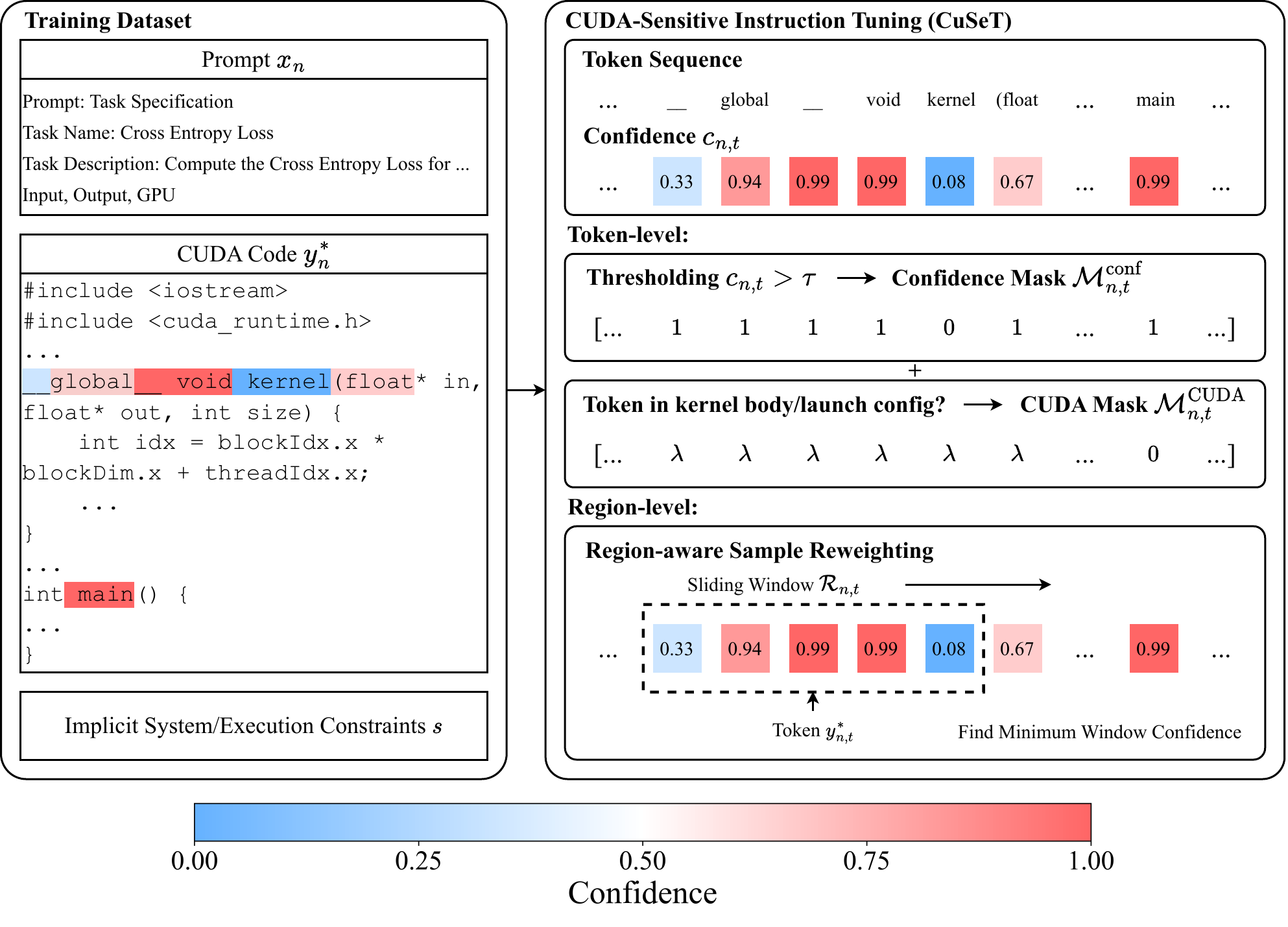}
    \caption{Overview of CuSeT, a CUDA-sensitive SFT framework. For each target token, CuSeT computes token confidence and derives a confidence mask, which is combined with a CUDA mask to form a hybrid token-level mask for execution-critical tokens. Sliding window analysis identifies low-confidence CUDA-sensitive regions, enabling sample reweighting. This pipeline focuses supervision on CUDA-sensitive tokens and regions critical for functional correctness.}
    \label{fig:illustration}
\end{figure}

Fig. \ref{fig:illustration} illustrates the overall architecture of CuSeT. Consider a dataset $\mathcal{D}=\{(x_n, y_n^*)\}_{n=1}^{N}$ consisting of $N$ prompt-response pairs, where each $x_n$ is the input prompt and $y_n^* = (y_{n,1}^*, y_{n,2}^*, \ldots, y_{n,T_n}^*)$ is the target CUDA code. The goal is to generate $y^*$ from $x$ using a policy $\pi_\theta$ parameterized by $\theta$, under implicit system and execution constraints $s$, including memory access patterns, thread synchronization, and indexing. CuSeT extends standard SFT by explicitly modeling CUDA-sensitive structures from tokens to regions, via adaptive token-level masking and region-aware sample reweighting.

\subsection{Adaptive Token-level Masking}
\label{subsec:token}
\paragraph{Confidence Mask.} 
Motivated by the confidence distribution analysis in Sec. \ref{sec:motivation}, we use token confidence as a dynamic signal to guide token-level supervision. For the $t$-th token $y_{n,t}^*$ in the $n$-th sample, its confidence is defined as the model's next-token prediction probability:
\begin{equation}
    c_{n,t}=\pi_\theta(y_{n,t}^* \mid x_n, y^*_{n,<t}),
\end{equation}
Using this confidence measure, we define a confidence mask $\mathcal{M}_{n,t}^{\text{conf}}$ as a binary indicator:
\begin{equation}
    \mathcal{M}_{n,t}^{\text{conf}}=\mathbbm{1}\left[\mathrm{sg}\left(\pi_\theta(y_{n,t}^* \mid x_n, y^*_{n,<t})>\tau\right)\right],
\end{equation}
where $\mathbbm{1}\left[\cdot\right]$ is the indicator function, $\mathrm{sg}(\cdot)$ denotes stop-gradient, and $\tau\in[0,1]$ is a static threshold. During training, the confidence mask selects high-confidence tokens for supervision while suppressing low-confidence tokens, effectively isolating parameter updates from less informative or noisy tokens. This also aligns with the probabilistic interpretation that low-probability tokens contribute smaller likelihood gradients \citep{liu2026profit}.

\paragraph{CUDA Mask.}
While the confidence mask preserves high-confidence tokens, it may suppress low-confidence CUDA-sensitive tokens that are nevertheless essential for correct CUDA execution. To preserve supervision for such tokens, we introduce a CUDA mask:
\begin{equation}
    \mathcal{M}_{n,t}^{\text{CUDA}}=\lambda \cdot \mathbbm{1}\{y_{n,t}^*\in\text{kernel body or launch config tokens of sample } n\},
\end{equation}
where $\lambda>0$ controls the weight assigned to tokens in execution-critical code locations. This mask ensures that tokens in kernel bodies and launch configurations retain a basic level of supervision even when their confidence is low.

\paragraph{Hybrid Mask.}
We construct a hybrid token-level mask by combining the confidence mask and the CUDA mask: 
\begin{equation}
    \mathcal{M}_{n,t}^{\text{hybrid}}=\mathcal{M}_{n,t}^{\text{conf}}+\mathcal{M}_{n,t}^{\text{CUDA}}.
\end{equation}
The hybrid mask selects high-confidence tokens through the confidence mask while retaining low-confidence tokens in execution-critical code locations through the CUDA mask. Meanwhile, low-confidence CUDA-neutral tokens receive reduced supervision. In this way, $\mathcal{M}_{n,t}^{\text{hybrid}}$ incorporates CUDA sensitivity into token-level supervision without treating all target tokens uniformly.

\subsection{Region-aware Sample Reweighting}
While the hybrid mask strengthens token-level supervision, it does not explicitly account for low-confidence CUDA-sensitive regions. As observed in Sec.~\ref{sec:motivation}, a small subset of CUDA-sensitive tokens forms low-confidence regions corresponding to execution-critical structures, such as shared memory allocation, which are critical for compilation and correct execution. Fig.~\ref{fig:confidence_curve} shows the sliding window confidence curve of a representative CUDA kernel, and Fig.~\ref{fig:zoom_in_seq_p} illustrates a low-confidence window corresponding to shared memory CUDA code. 

These observations indicate that supervision should extend beyond individual tokens to the region level, since low-confidence regions often correspond to CUDA-sensitive code structures that encode complex execution constraints spanning multiple tokens. To incorporate this structure, we propose region-aware sample reweighting based on sliding window confidence analysis, which assigns higher weights to samples containing such low-confidence CUDA-sensitive regions. Specifically, for each token $y_{n,t}^*$, following the design principles in \citep{fu2025deep,tang2025aligning}, we associate it with a sliding window $\mathcal{R}_{n,t}$ of size $l$ centered at position $t$:
\begin{equation}
    \mathcal{R}_{n,t}=\left\{y_{n,j}^* \; \middle| \;\max\left(1, t-\left\lfloor\frac{l}{2}\right\rfloor\right)\leq j \leq\min\left(T_n, t+\left\lceil\frac{l}{2}\right\rceil-1\right)
    \right\},
\end{equation}
The window is clipped to valid token positions at sample sequence boundaries, and if $T_n < l$, it covers the entire sequence. We define the window confidence as the average token confidence within each sliding window:
\begin{equation}
    C_{n,t}=\frac{1}{|\mathcal{R}_{n,t}|}\sum_{y_{n,j}^*\in \mathcal{R}_{n,t}}c_{n,j}=\frac{1}{|\mathcal{R}_{n,t}|}\sum_{y_{n,j}^*\in \mathcal{R}_{n,t}}\pi_\theta(y_{n,j}^* \mid x_n, y^*_{n,<j}),
\end{equation}
For each sample, the most difficult region is identified by the minimum window confidence:
\begin{equation}
    \mathcal{C}_{n}=\min_{1\leq t\leq T_n}\{C_{n,t}\},
\end{equation}
The sample-level weight is then defined as:
\begin{equation}
    w_n=\mathrm{sg}\left(1-\mathcal{C}_n\right). 
\end{equation}
This assigns higher weights to samples containing low-confidence CUDA-sensitive regions, encouraging the model to focus on challenging CUDA-sensitive structures during optimization.

\subsection{Training Objective}
We combine the token-level hybrid mask and the sample-level weight to define the final CuSeT objective. The model is trained with a likelihood objective that incorporates adaptive token-level masking and region-aware sample reweighting:
\begin{equation}
    \mathcal{L}_{\text{CuSeT}}(\theta) = \mathbb{E}_{\{(x_n,y_n^*,s)\}_{n=1}^{N}} \left[ - w_n \sum_{t=1}^{T_n} \mathcal{M}_{n,t}^{\text{hybrid}}\,\log \pi_\theta(y_{n,t}^* \mid x_{n}, y^*_{n,<t}) \right].
\end{equation}
This objective implements CUDA-sensitive supervision from tokens to regions. At the token level, $\mathcal{M}_{n,t}^{\text{hybrid}}$ preserves high-confidence tokens while retaining low-confidence CUDA-sensitive tokens in execution-critical code locations. At the sample level, $w_n$ increases the training emphasis on samples containing low-confidence regions identified via sliding window analysis. Overall, CuSeT strengthens supervision on execution-critical tokens while prioritizing samples containing low-confidence CUDA-sensitive regions, thereby enhancing the model's focus on CUDA-sensitive structures and improving functional correctness.

\section{Evaluation}
\subsection{Experimental Setup}
\paragraph{Data Gathering Pipeline and Benchmarks.} We construct a diverse CUDA kernel dataset from multiple sources, including open-source repositories such as GitHub and Hugging Face, as well as benchmark datasets \citep{gong2025large,wen2022babeltower} and PyTorch modules \citep{ouyang2025kernelbench,kernelbook2025}. To ensure functional correctness, we use an LLM (e.g., Qwen3-Coder \citep{yang2025qwen3}) to generate five sets of input cases for each kernel and derive corresponding reference outputs for validation. Each kernel is compiled and executed on a GPU, and only those passing both compilation and functional correctness checks are retained. This results in a dataset of 6,278 high-quality prompt–response pairs covering diverse CUDA kernel patterns and execution constraints. For evaluation, we use the CUDABench Level 1 benchmark \citep{zhu2026cudabench}, which covers multiple application domains and input scales to assess CUDA kernel generation.

\paragraph{Baselines and Metrics.} We compare CuSeT
against standard SFT. Additionally, we include several improved SFT methods, including DFT \citep{wu2025generalization}, EAFT \citep{diao2026entropy}, ASFT \citep{zhu2025anchored}, and ProFit \citep{liu2026profit}. We evaluate the generated kernels using \textbf{Compilation Success} and \textbf{Functional Correctness}. We report results using \textbf{Pass@k} ($k=\{1, 3\}$) \citep{chen2021evaluating}, which estimates the probability that at least one valid CUDA kernel is generated within $k$ sampled outputs.

\paragraph{Models and Implementation Details.}
To verify the generalizability of CuSeT across model families, we conduct experiments on a diverse set of code LLMs, including Qwen2.5-Coder-7B-Instruct \citep{hui2024qwen2}, DeepSeek-Coder-6.7B-Instruct \citep{guo2024deepseek}, and Seed-Coder-8B-Instruct \citep{seed2025seed}. In the training stage, we utilize LLaMAFactory \citep{zheng2024llamafactory} framework and perform LoRA \citep{hulora} fine-tuning with LoRA rank $r=8$, $\alpha=16$, and the LoRA dropout rate 0.05 for all linear layers. The maximum sequence length is set to 4,096, with a batch size of 1 per GPU. The model is fine-tuned with a learning rate of $1\times 10^{-4}$ for 3 epochs on 8 NVIDIA RTX 3090 Ti GPUs. For CuSeT specific hyperparameters, in adaptive token-level masking, we set the confidence mask threshold to $\tau=0.1$ and the CUDA mask scaling factor to $\lambda=1.0$, ensuring that tokens in kernel bodies and launch configurations receive basic supervision even under low confidence. In region-aware sample reweighting, we use a sliding window of size $l=256$ to compute window confidence.

\subsection{Performance Analysis}
The main results are reported in Tab. \ref{tab:main_res}. CuSeT consistently outperforms standard SFT and representative improved SFT methods across all model families in terms of functional correctness, while maintaining consistently high compilation success.

\begin{table}[!ht]
\centering
\begin{footnotesize}
\centering
\caption{Main results on CUDABench Level 1 benchmark. We report the accuracy of the vanilla baseline, standard SFT, and varying improved SFT strategies (DFT, EAFT, ASFT, ProFit) on multiple model families. Best results are \textbf{bolded}, and second-best are \underline{underlined}.}
\begin{tabular}{cccccc}
\toprule
\multirow{2}{*}{\textbf{Model}}                        & \multirow{2}{*}{\textbf{Method}} & \multicolumn{2}{c}{\textbf{Pass@1}} & \multicolumn{2}{c}{\textbf{Pass@3}} \\ \cmidrule(lr){3-4} \cmidrule(l){5-6} 
                                              &                         & \textbf{Compilation}   & \textbf{Function}   & \textbf{Compilation}   & \textbf{Function}   \\ \midrule
\multirow{7}{*}{Qwen2.5-Coder-7B-Instruct}    & Vanilla                 & 69.0            & 35.2       & 83.6          & 49.2       \\
                                              & SFT                     & 92.2          & 55.0         & 96.6          & 61.0         \\
                                              & DFT                     & 93.0            & \underline{60.2}       & 93.8          & 62.0         \\
                                              & EAFT                    & 91.4          & 56.8       & \underline{97.2}          & \underline{65.2}       \\
                                              & ASFT                    & 77.8          & 50.4       & 81.6          & 54.4       \\
                                              & ProFit                  & \textbf{94.2}          & 59.6       & 97.0            & 64.0         \\
                                              \rowcolor{blue!5} \cellcolor{white} & CuSeT                  & \underline{93.6}            & \textbf{65.6}       & \textbf{98.8}            & \textbf{74.8}         \\ \midrule
\multirow{7}{*}{DeepSeek-Coder-6.7B-Instruct} & Vanilla                 & 80.2          & 42.4       & 87.8          & 48.4       \\
                                              & SFT                     & 90.4          & 54.6       & 96.2          & 62.2       \\
                                              & DFT                     & 88.0            & 53.0         & 90.6          & 55.0         \\
                                              & EAFT                    & 89.4          & 53.4       & 94.0            & 60.2       \\
                                              & ASFT                    & 89.4          & 51.8       & 93.2          & 56.0         \\
                                              & ProFit                  & \underline{91.8}          & \underline{56.8}       & \underline{97.0}            & \underline{63.6}       \\
                                              \rowcolor{blue!5} \cellcolor{white} & CuSeT                  & \textbf{95.0}          & \textbf{63.2}       & \textbf{99.0}            & \textbf{70.6}       \\ \midrule
\multirow{7}{*}{Seed-Coder-8B-Instruct}       & Vanilla                 & 4.0             & 2.8        & 9.0             & 6.6        \\
                                              & SFT                     & 93.4          & 61.2       & 97.6          & 68.8       \\
                                              & DFT                     & \underline{97.0}            & 66.4       & 98.0            & 67.6       \\
                                              & EAFT                    & 94.6          & 62.6       & 99.2          & \underline{69.8}       \\
                                              & ASFT                    & 88.8          & 59.0         & 91.4          & 64.8       \\
                                              & ProFit                  & \textbf{97.6}          & \underline{67.6}       & \underline{98.4}          & 69.6       \\
                                              \rowcolor{blue!5} \cellcolor{white} & CuSeT                  & 94.6          & \textbf{67.8}       & \textbf{98.8}          & \textbf{75.8}         \\ \bottomrule
\end{tabular}
\label{tab:main_res}
\end{footnotesize}
\end{table}

For instance, in Qwen2.5-Coder-7B-Instruct, CuSeT achieves 65.6\% functional correctness under Pass@1, exceeding the best baseline by over 5\%, while maintaining high compilation success under Pass@1. Similar patterns are observed for DeepSeek-Coder-6.7B-Instruct and Seed-Coder-8B-Instruct, where CuSeT consistently improves functional correctness up to 6.4\% over the strongest baselines. These results indicate that combining adaptive token-level masking and region-aware sample reweighting improves the model's ability to capture CUDA-sensitive structures, resulting in higher functional correctness in CUDA kernel generation.

\subsection{Comparison with SOTA Methods}
We compare CuSeT-tuned models against two groups of baselines for CUDA kernel generation, including frontier general-purpose and code models such as GLM-5.1 \citep{glm51}, DeepSeek-V3.2-Thinking \citep{liu2025deepseek} and Qwen3-Coder-Next \citep{cao2026qwen3}, as well as CUDA-specialized models such as Kevin \citep{baronio2025kevin} and KernelCoder \citep{kong2025concur}. Tab.~\ref{tab:sota} summarizes the results on CUDABench Level 1 benchmark.

\begin{table}[!ht]
\centering
\setlength{\tabcolsep}{1.5pt}
\begin{footnotesize}
\centering
\caption{Comparison of frontier models and CuSeT-tuned models on the CUDABench Level 1 benchmark. Avg. Consumed Tokens denotes the average number of tokens generated per sample during inference, serving as a measure of generation efficiency.}
\begin{tabular}{cccccccc}
\toprule
\multirow{2}{*}{\textbf{Model}}              & \multirow{2}{*}{\textbf{Type}} & \multirow{2}{*}{\textbf{\#Params}} & \multicolumn{2}{c}{\textbf{Pass@1}} & \multicolumn{2}{c}{\textbf{Pass@3}} & \multirow{2}{*}{\begin{tabular}[c]{@{}c@{}}\textbf{Avg. Consumed}\\\textbf{Tokens}\end{tabular}} \\ \cmidrule(lr){4-5} \cmidrule(lr){6-7}
                                    &                       &                           & \textbf{Compilation}   & \textbf{Function}   & \textbf{Compilation}   & \textbf{Function}   &                              \\ \midrule
\multicolumn{8}{l}{Frontier Models}                                                                                                                                              \\ \midrule
GLM-5.1                            & Reasoning             & 40B/744B                        & 16.8          & 15.6       & 32.2          & 29.2       & 8,220                             \\
DeepSeek-V3.2-Thinking                            & Reasoning             & 685B                        & 93.8          & 66.6       & 99.4          & 80.6       & 3,717                             \\
Qwen3-Coder-Next                           & Code             & 80B                        & 86.0          & 63.4       & 98.0          & 80.8       &  1,273                            \\
Kevin                               & Reasoning             & 32B                       & 96.0          & 68.8       & \textbf{99.6}          & 78.2       & 5,392                             \\
KernelCoder                        & Reasoning             & 32B                       & \textbf{96.8}          & 70.6       & \textbf{99.6}          & 79.8       & 5,379                             \\ \midrule
\multicolumn{8}{l}{Models Fine-tuned with CuSeT}                                                                                                                                         \\ \midrule
Qwen2.5-Coder-7B-Instruct    & Code                  & 7B                        & 93.6          & 65.6       & 98.8          & 74.8       & 743                             \\
DeepSeek-Coder-6.7B-Instruct & Code                  & 6.7B                      & 95.0            & 63.2       & 99.0            & 70.6       & 1,018                             \\
Seed-Coder-8B-Instruct       & Code                  & 8B                        & 94.6          & 67.8       & 98.8          & 75.8       & 839                             \\
Qwen3-14B                   & General               & 14B                        & 92.6              & 66.8           & 97.6              & 76.0           & \textbf{722}                             \\ 
Qwen2.5-Coder-32B-Instruct    & Code                  & 32B                        & 96.0          & \textbf{75.6}       & 98.8          & \textbf{81.0}       & 726                             \\ \bottomrule
\end{tabular}
\label{tab:sota}
\end{footnotesize}
\end{table}

CuSeT-tuned models consistently achieve competitive or superior functional correctness while significantly reducing inference cost measured by average generated tokens. In particular, Qwen2.5-Coder-32B-Instruct with CuSeT achieves 75.6\% (Pass@1) and 81.0\% (Pass@3) functional correctness, while using only 726 tokens on average, substantially fewer than frontier baselines. These results demonstrate a favorable trade-off between generation quality and efficiency, indicating that explicitly modeling CUDA-sensitive tokens and regions enables more efficient kernel generation without relying on costly multi-stage training or long-context reasoning pipelines.

\subsection{Scalability}
To evaluate the scalability of CuSeT, we conduct experiments on models of different sizes, including Qwen3-4B \citep{yang2025qwen3} and Qwen2.5-Coder-32B-Instruct \citep{hui2024qwen2}. Tab.~\ref{tab:scale} reports the results compared with the vanilla pretrained models, standard SFT, and CuSeT.
\begin{table}[!ht]
\centering
\begin{footnotesize}
\centering
\caption{Scalability evaluation of CuSeT across models of different sizes.}
\begin{tabular}{lcccc}
\toprule
\multicolumn{1}{c}{\multirow{2}{*}{\textbf{Method}}} & \multicolumn{2}{c}{\textbf{Pass@1}} & \multicolumn{2}{c}{\textbf{Pass@3}} \\ \cmidrule(lr){2-3} \cmidrule(lr){4-5} 
\multicolumn{1}{c}{} & \textbf{Compilation}   & \textbf{Function}   & \textbf{Compilation}   & \textbf{Function}   \\ \midrule
Qwen3-4B      & 70.6          & 23.0       & 82.2          & 33.6       \\
+SFT                & 89.6          & 51.2       & \textbf{98.4}          & 65.0       \\
+CuSeT                        & \textbf{91.6}           & \textbf{57.8}       & 97.6          & \textbf{66.2}       \\ \midrule
Qwen2.5-Coder-32B-Instruct              & 92.6          & 60.2       & 97.6          & 70.0       \\
+SFT                & 94.4          & 68.0       & 98.6          & 79.6       \\
+CuSeT                        & \textbf{96.0}          & \textbf{75.6}       & \textbf{98.8}          & \textbf{81.0}       \\ \bottomrule
\end{tabular}
\label{tab:scale}
\end{footnotesize}
\end{table}
CuSeT consistently improves functional correctness over standard SFT across both small and large models, demonstrating that its effectiveness generalizes across model scales and architectures, indicating that it effectively captures CUDA-sensitive code structures. These results suggest that CuSeT is scalable and robust, enabling consistent improvements without additional training complexity or data overhead.

\subsection{Ablation Study}
\subsubsection{Component Analysis}
We conduct ablation studies in Tab.~\ref{tab:ablation} to evaluate each component of CuSeT. The results show consistent improvements from each design choice.
\begin{table}[!ht]
\centering
\setlength{\tabcolsep}{5pt}
\begin{footnotesize}
\centering
\caption{Ablation study of CuSeT, evaluating the contribution of the confidence mask, CUDA mask, and region-aware sample reweighting.}
\begin{tabular}{ccccccc}
\toprule
\multirow{2}{*}{\textbf{Confidence Mask}} & \multirow{2}{*}{\textbf{CUDA Mask}} & \multirow{2}{*}{\begin{tabular}[c]{@{}c@{}}\textbf{Region-aware}\\\textbf{Sample Reweighting}\end{tabular}} & \multicolumn{2}{c}{\textbf{Pass@1}} & \multicolumn{2}{c}{\textbf{Pass@3}} \\ \cmidrule(lr){4-5} \cmidrule(lr){6-7} 
                                 &                            &                               & \textbf{Compilation}   & \textbf{Function}   & \textbf{Compilation}   & \textbf{Function}   \\ \midrule
\ding{56}                                & \ding{56}                           & \ding{56}                              & 92.2          & 55.0         & 96.6          & 61.0       \\
\ding{52}                                & \ding{56}                           & \ding{56}                              & 94.2          & 59.6       & 97.0          & 64.0       \\
\ding{52}                                & \ding{52}                           &  \ding{56}                             & 96.6              & 62.8           & 99.4              & 70.4         \\
\ding{52}                                 & \ding{56}                           & \ding{52}                              & 94.6              & 61.8           & 98.0              & 68.0           \\
\ding{52}                                 & \ding{52}                          & \ding{52}                              & 93.6          & 65.6       & 98.8          & 74.8       \\ \bottomrule
\end{tabular}
\label{tab:ablation}
\end{footnotesize}
\end{table}
(1) Introducing the confidence mask improves functional correctness over standard SFT, indicating that token-level confidence-based reweighting enhances supervision by focusing on high-confidence CUDA-sensitive tokens. (2) Adding the CUDA mask further improves functional correctness, demonstrating that explicitly preserving supervision for execution-critical tokens in kernel bodies and launch configurations provides complementary gains beyond confidence-based reweighting. (3) Incorporating region-aware sample reweighting leads to additional improvements in functional correctness, suggesting that emphasizing samples containing low-confidence CUDA-sensitive regions helps the model better learn difficult execution-critical patterns. (4) Combining all components yields the best overall functional correctness, confirming that token-level and region-level CUDA sensitivity modeling are complementary and jointly improve CUDA kernel generation performance.

\subsubsection{Hyper-parameter Analysis}
\begin{figure}[!ht]
    \centering
    \begin{subfigure}[t]{0.33\linewidth}
        \centering
        \includegraphics[width=\textwidth]{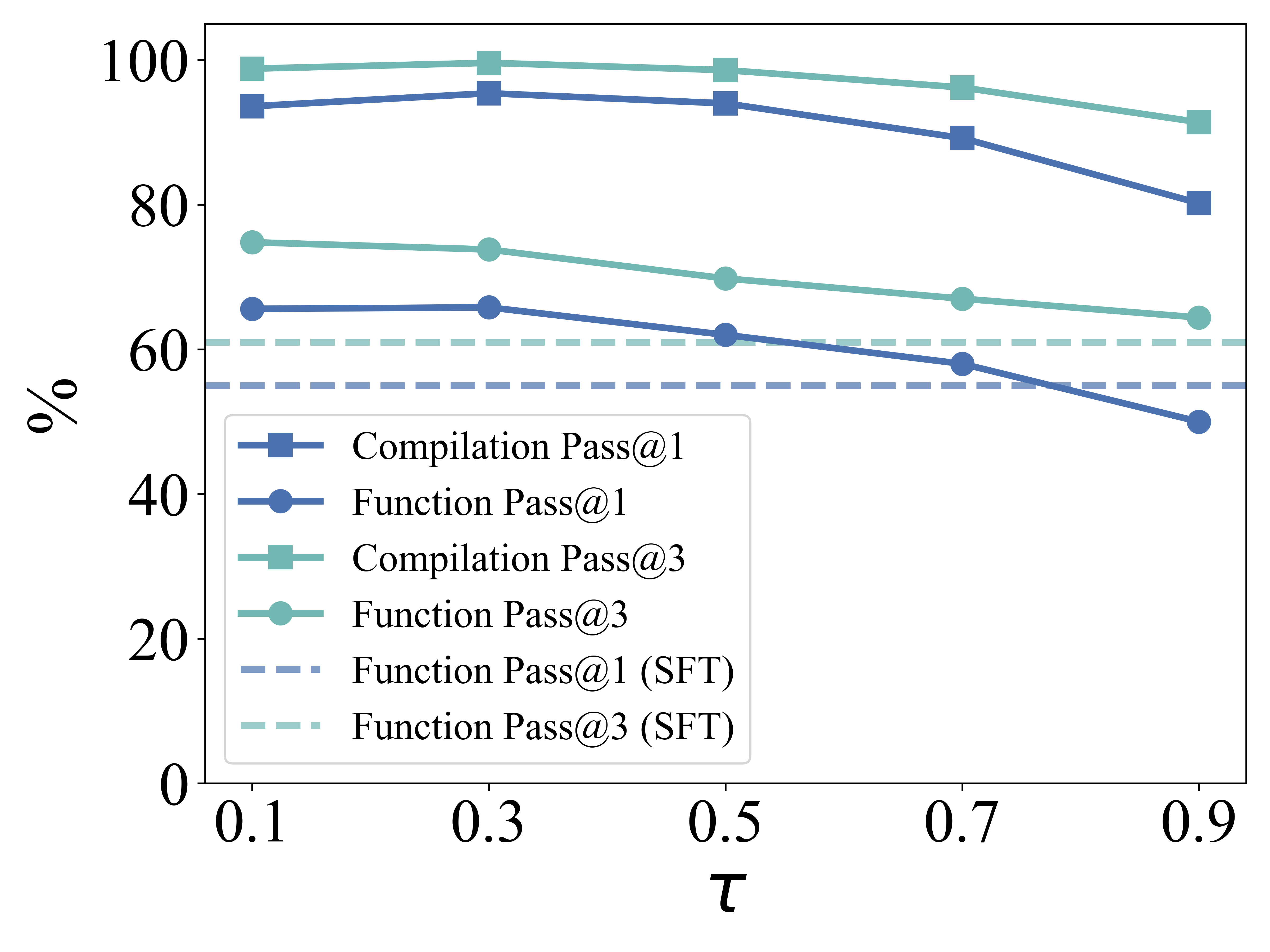}
        \caption{Confidence Mask Threshold $\tau$}
        \label{fig:tau}
    \end{subfigure}\hfill
    \begin{subfigure}[t]{0.33\linewidth}
        \centering
        \includegraphics[width=\textwidth]{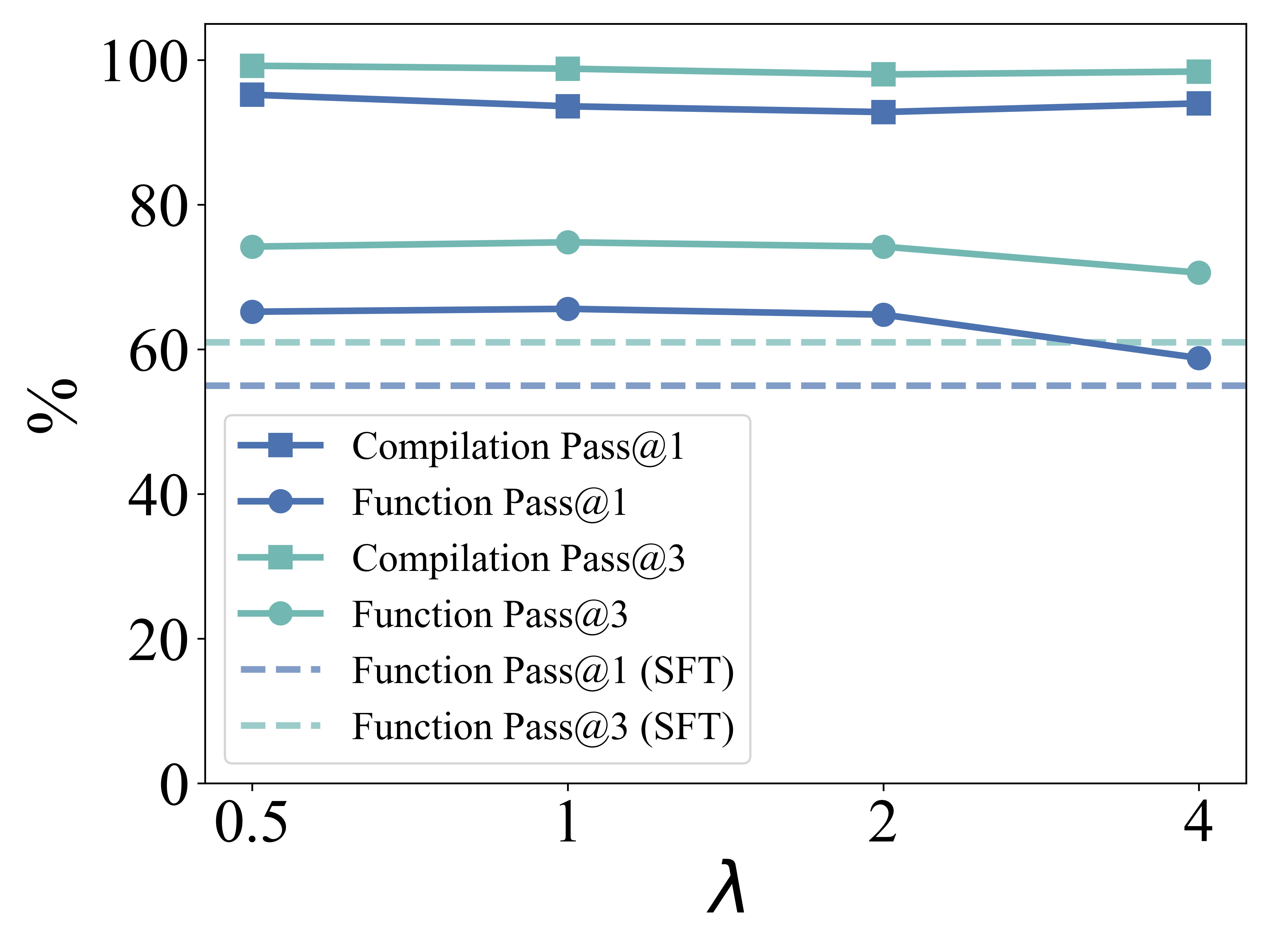}
        \caption{CUDA Mask Scale $\lambda$}
        \label{fig:lambda}
    \end{subfigure}\hfill
    \begin{subfigure}[t]{0.33\linewidth}
        \centering
        \includegraphics[width=\textwidth]{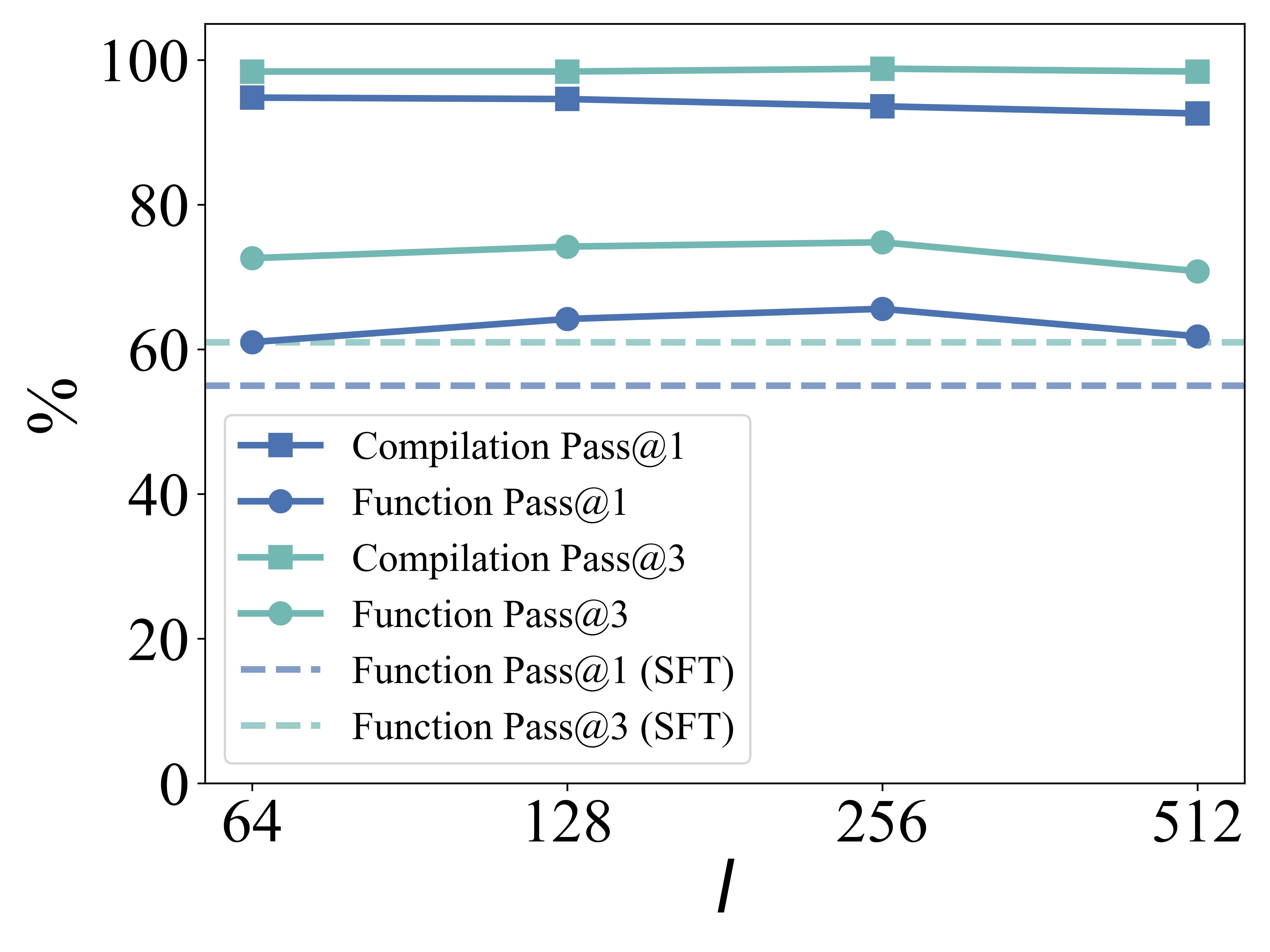}
        \caption{Sliding Window Size $l$}
        \label{fig:l}
    \end{subfigure}
    \caption{Ablation analysis of hyper-parameters.}
    \label{fig:}
\end{figure}

We further analyze the sensitivity of CuSeT to hyper-parameters. All experiments are conducted on Qwen2.5-Coder-7B-Instruct to ensure fair comparison.
\paragraph{Analysis of Confidence Mask Threshold $\tau$.} To study the effect of the confidence mask threshold $\tau$, we vary it from 0.1 to 0.9 and report results in Fig. \ref{fig:tau}. We observe that relatively low thresholds (e.g., $\tau=0.1$ and $\tau=0.3$) achieve the best overall performance. This indicates that a moderate confidence mask effectively removes low-confidence CUDA-neutral tokens while preserving CUDA-sensitive tokens. In contrast, larger thresholds lead to consistent performance degradation, as overly strict filtering removes important CUDA-sensitive tokens with lower confidence. These results suggest that a relatively lower threshold provides a better balance between suppressing low-confidence CUDA-neutral tokens and retaining CUDA-sensitive tokens.

\paragraph{Analysis of CUDA Mask Scale $\lambda$.}
We further study the effect of the CUDA mask scale $\lambda$, which controls the supervision strength applied to kernel bodies and launch configurations. As shown in Fig.~\ref{fig:lambda}, a moderate value achieves the best performance, with $\lambda=1$ yielding the highest functional correctness under both Pass@1 (65.6\%) and Pass@3 (74.8\%). When $\lambda$ is too small, the CUDA mask provides insufficient additional supervision, limiting its ability to preserve execution-critical information. In contrast, overly large values overemphasize execution-critical code location and disrupt the balance between general code modeling and CUDA sensitivity learning. For example, increasing $\lambda$ to 4 reduces Pass@1 functional correctness from 65.6\% to 58.8\%. These results suggest that a properly balanced CUDA mask scale is important for effectively enhancing supervision on execution-critical code locations while maintaining overall modeling stability.

\paragraph{Analysis of Sliding Window Size $l$.}
We further study the impact of the sliding window size $l$ used for computing window confidence. As shown in Fig.~\ref{fig:l}, the performance is sensitive to the choice of $l$. When the window size is too small (e.g., $l=64$), the computed confidence mainly reflects short-range token fluctuations and provides a limited view of low-confidence CUDA-sensitive regions, leading to lower functional correctness. Increasing the window size to $l=128$ and $l=256$ improves performance, indicating that a moderate window is better at capturing execution-critical patterns in low-confidence CUDA-sensitive regions, such as shared memory allocation, indexing, and launch configuration. However, overly large windows degrade performance. When $l=512$, Pass@1 functional correctness drops from 65.6\% to 61.8\%, suggesting that large windows over-smooth the confidence signal and mix low-confidence CUDA-sensitive regions with surrounding CUDA-neutral code. Overall, $l=256$ achieves the best performance, suggesting that CuSeT benefits from a window size that is sufficiently large to capture execution-critical patterns at the region level, while avoiding excessive interference from surrounding CUDA-neutral code.

\section{Conclusion}
In this work, we present \textbf{CUDA-Sensitive Instruction Tuning (CuSeT)}, a low-cost post-training method for improving LLM-based CUDA kernel generation. Motivated by the observation that CUDA sensitivity appears at both token and region levels, CuSeT enhances supervision over CUDA-sensitive structures within a simple SFT framework. Specifically, it integrates token-level confidence with a CUDA mask to preserve supervision on execution-critical tokens while maintaining coverage of critical code locations. At the same time, region-aware sample reweighting emphasizes training on samples containing low-confidence CUDA-sensitive regions. Extensive experiments across multiple model families and scales demonstrate that CuSeT consistently improves functional correctness while maintaining high compilation success. CuSeT outperforms standard SFT and improved SFT variants, achieves competitive performance against frontier CUDA kernel generation models, and significantly reduces inference token cost. Overall, CuSeT provides a simple yet effective way to incorporate CUDA sensitivity into post-training, enabling LLMs to better capture execution-constrained code structures and generate more reliable CUDA kernels.

\bibliographystyle{unsrt}
\bibliography{custom}

@article{ouyang2025kernelbench,
  title={Kernelbench: Can llms write efficient gpu kernels?},
  author={Ouyang, Anne and Guo, Simon and Arora, Simran and Zhang, Alex L and Hu, William and R{\'e}, Christopher and Mirhoseini, Azalia},
  journal={arXiv preprint arXiv:2502.10517},
  year={2025}
}

@article{li2025autotriton,
  title={Autotriton: Automatic triton programming with reinforcement learning in llms},
  author={Li, Shangzhan and Wang, Zefan and He, Ye and Li, Yuxuan and Shi, Qi and Li, Jianling and Hu, Yonggang and Che, Wanxiang and Han, Xu and Liu, Zhiyuan and others},
  journal={arXiv preprint arXiv:2507.05687},
  year={2025}
}

@article{wei2025astra,
  title={Astra: A Multi-Agent System for GPU Kernel Performance Optimization},
  author={Wei, Anjiang and Sun, Tianran and Seenichamy, Yogesh and Song, Hang and Ouyang, Anne and Mirhoseini, Azalia and Wang, Ke and Aiken, Alex},
  journal={arXiv preprint arXiv:2509.07506},
  year={2025}
}

@article{chen2025cuda,
  title={CUDA-LLM: LLMs Can Write Efficient CUDA Kernels},
  author={Chen, Wentao and Zhu, Jiace and Fan, Qi and Ma, Yehan and Zou, An},
  journal={arXiv preprint arXiv:2506.09092},
  year={2025}
}

@article{kong2025concur,
  title={ConCuR: Conciseness Makes State-of-the-Art Kernel Generation},
  author={Kong, Lingcheng and Wei, Jiateng and Shen, Hanzhang and Wang, Huan},
  journal={arXiv preprint arXiv:2510.07356},
  year={2025}
}

@techreport{lange2025ai,
  title={The AI CUDA engineer: Agentic CUDA kernel discovery, optimization and composition},
  author={Lange, Robert Tjarko and Prasad, Aaditya and Sun, Qi and Faldor, Maxence and Tang, Yujin and Ha, David},
  year={2025},
  institution={Technical report, Sakana AI, 02 2025}
}

@article{dong2025stark,
  title={STARK: Strategic Team of Agents for Refining Kernels},
  author={Dong, Juncheng and Yang, Yang and Liu, Tao and Wang, Yang and Qi, Feng and Tarokh, Vahid and Rangadurai, Kaushik and Yang, Shuang},
  journal={arXiv preprint arXiv:2510.16996},
  year={2025}
}

@article{ouyang2022training,
  title={Training language models to follow instructions with human feedback},
  author={Ouyang, Long and Wu, Jeffrey and Jiang, Xu and Almeida, Diogo and Wainwright, Carroll and Mishkin, Pamela and Zhang, Chong and Agarwal, Sandhini and Slama, Katarina and Ray, Alex and others},
  journal={Advances in neural information processing systems},
  volume={35},
  pages={27730--27744},
  year={2022}
}

@article{baronio2025kevin,
  title={Kevin: Multi-turn rl for generating cuda kernels},
  author={Baronio, Carlo and Marsella, Pietro and Pan, Ben and Guo, Simon and Alberti, Silas},
  journal={arXiv preprint arXiv:2507.11948},
  year={2025}
}

@inproceedings{wei2024magicoder,
  title={Magicoder: empowering code generation with OSS-INSTRUCT},
  author={Wei, Yuxiang and Wang, Zhe and Liu, Jiawei and Ding, Yifeng and Zhang, Lingming},
  booktitle={Proceedings of the 41st International Conference on Machine Learning},
  pages={52632--52657},
  year={2024}
}

@software{kernelbook2025,
    title={KernelBook},
    author={Paliskara, Sahan and Saroufim, Mark},
    year={2025},
    month={5},
    url={https://huggingface.co/datasets/GPUMODE/KernelBook},
}

@article{lei2025pragma,
  title={PRAGMA: A Profiling-Reasoned Multi-Agent Framework for Automatic Kernel Optimization},
  author={Lei, Kelun and Yang, Hailong and Zhang, Huaitao and You, Xin and Zhang, Kaige and Luan, Zhongzhi and Liu, Yi and Qian, Depei},
  journal={arXiv preprint arXiv:2511.06345},
  year={2025}
}

@article{li2025cuda,
  title={Cuda-l1: Improving cuda optimization via contrastive reinforcement learning},
  author={Li, Xiaoya and Sun, Xiaofei and Wang, Albert and Li, Jiwei and Shum, Chris},
  journal={arXiv preprint arXiv:2507.14111},
  year={2025}
}

@article{gong2025large,
  title={From Large to Small: Transferring CUDA Optimization Expertise via Reasoning Graph},
  author={Gong, Junfeng and Wei, Zhiyi and Chen, Junying and Liu, Cheng and Li, Huawei},
  journal={arXiv preprint arXiv:2510.19873},
  year={2025}
}

@article{woo2025tritonrl,
  title={TritonRL: Training LLMs to Think and Code Triton Without Cheating},
  author={Woo, Jiin and Zhu, Shaowei and Nie, Allen and Jia, Zhen and Wang, Yida and Park, Youngsuk},
  journal={arXiv preprint arXiv:2510.17891},
  year={2025}
}

@article{seed2025seed,
  title={Seed-coder: Let the code model curate data for itself},
  author={Seed, ByteDance and Zhang, Yuyu and Su, Jing and Sun, Yifan and Xi, Chenguang and Xiao, Xia and Zheng, Shen and Zhang, Anxiang and Liu, Kaibo and Zan, Daoguang and others},
  journal={arXiv preprint arXiv:2506.03524},
  year={2025}
}

@inproceedings{zheng2024llamafactory,
  title={Llamafactory: Unified efficient fine-tuning of 100+ language models},
  author={Zheng, Yaowei and Zhang, Richong and Zhang, Junhao and Ye, Yanhan and Luo, Zheyan},
  booktitle={Proceedings of the 62nd annual meeting of the association for computational linguistics (volume 3: system demonstrations)},
  pages={400--410},
  year={2024}
}

@article{chen2021evaluating,
  title={Evaluating large language models trained on code},
  author={Chen, Mark},
  journal={arXiv preprint arXiv:2107.03374},
  year={2021}
}

@inproceedings{wen2022babeltower,
  title={Babeltower: Learning to auto-parallelized program translation},
  author={Wen, Yuanbo and Guo, Qi and Fu, Qiang and Li, Xiaqing and Xu, Jianxing and Tang, Yanlin and Zhao, Yongwei and Hu, Xing and Du, Zidong and Li, Ling and others},
  booktitle={International Conference on Machine Learning},
  pages={23685--23700},
  year={2022},
  organization={PMLR}
}

@article{zhu2026cudabench,
  title={CUDABench: Benchmarking LLMs for Text-to-CUDA Generation},
  author={Zhu, Jiace and Chen, Wentao and Fan, Qi and Ren, Zhixing and Wu, Junying and Chai, Xing Zhe and Rungrueangwutthinon, Chotiwit and Ma, Yehan and Zou, An},
  journal={arXiv preprint arXiv:2603.02236},
  year={2026}
}

@article{guo2024deepseek,
  title={DeepSeek-Coder: when the large language model meets programming--the rise of code intelligence},
  author={Guo, Daya and Zhu, Qihao and Yang, Dejian and Xie, Zhenda and Dong, Kai and Zhang, Wentao and Chen, Guanting and Bi, Xiao and Wu, Yifan and Li, YK and others},
  journal={arXiv preprint arXiv:2401.14196},
  year={2024}
}

@article{wu2025generalization,
  title={On the generalization of sft: A reinforcement learning perspective with reward rectification},
  author={Wu, Yongliang and Zhou, Yizhou and Ziheng, Zhou and Peng, Yingzhe and Ye, Xinyu and Hu, Xinting and Zhu, Wenbo and Qi, Lu and Yang, Ming-Hsuan and Yang, Xu},
  journal={arXiv preprint arXiv:2508.05629},
  year={2025}
}

@article{liu2026profit,
  title={ProFit: Leveraging High-Value Signals in SFT via Probability-Guided Token Selection},
  author={Liu, Tao and Wu, Taiqiang and Yang, Runming and Sun, Shaoning and Wang, Junjie and Yang, Yujiu},
  journal={arXiv preprint arXiv:2601.09195},
  year={2026}
}

@article{hui2024qwen2,
  title={Qwen2. 5-coder technical report},
  author={Hui, Binyuan and Yang, Jian and Cui, Zeyu and Yang, Jiaxi and Liu, Dayiheng and Zhang, Lei and Liu, Tianyu and Zhang, Jiajun and Yu, Bowen and Lu, Keming and others},
  journal={arXiv preprint arXiv:2409.12186},
  year={2024}
}

@article{diao2026entropy,
  title={Entropy-Adaptive Fine-Tuning: Resolving Confident Conflicts to Mitigate Forgetting},
  author={Diao, Muxi and Yang, Lele and Gong, Wuxuan and Zhang, Yutong and Yan, Zhonghao and Han, Yufei and Liang, Kongming and Xu, Weiran and Ma, Zhanyu},
  journal={arXiv preprint arXiv:2601.02151},
  year={2026}
}

@article{zhu2025anchored,
  title={Anchored Supervised Fine-Tuning},
  author={Zhu, He and Su, Junyou and Lai, Peng and Ma, Ren and Zhang, Wenjia and Yang, Linyi and Chen, Guanhua},
  journal={arXiv preprint arXiv:2509.23753},
  year={2025}
}

@article{dao2022flashattention,
  title={Flashattention: Fast and memory-efficient exact attention with io-awareness},
  author={Dao, Tri and Fu, Dan and Ermon, Stefano and Rudra, Atri and R{\'e}, Christopher},
  journal={Advances in neural information processing systems},
  volume={35},
  pages={16344--16359},
  year={2022}
}

@article{saba2026cutegen,
  title={CuTeGen: An LLM-Based Agentic Framework for Generation and Optimization of High-Performance GPU Kernels using CuTe},
  author={Saba, Tara and Ouyang, Anne and Si, Xujie and Long, Fan},
  journal={arXiv preprint arXiv:2604.01489},
  year={2026}
}

@software{kernelllm2025,
    title={KernelLLM: Making Kernel Development More Accessible},
    author={Fisches, Zacharias V. and Paliskara, Sahan and Guo, Simon and Zhang, Alex and Spisak, Joe and Cummins, Chris and Leather, Hugh and Synnaeve, Gabriel and Isaacson, Joe and Markosyan, Aram and Saroufim, Mark},
    year={2025},
    month={6},
    url={https://huggingface.co/facebook/KernelLLM},
}

@article{lv2025hpctranscompile,
  title={Hpctranscompile: An ai compiler generated dataset for high-performance cuda transpilation and llm preliminary exploration},
  author={Lv, Jiaqi and He, Xufeng and Liu, Yanchen and Dai, Xu and Shen, Aocheng and Li, Yinghao and Hao, Jiachen and Ding, Jianrong and Hu, Yang and Yin, Shouyi},
  journal={arXiv preprint arXiv:2506.10401},
  year={2025}
}

@book{cook2012cuda,
  title={CUDA programming: a developer's guide to parallel computing with GPUs},
  author={Cook, Shane},
  year={2012},
  publisher={Newnes}
}

@article{roziere2023code,
  title={Code llama: Open foundation models for code},
  author={Roziere, Baptiste and Gehring, Jonas and Gloeckle, Fabian and Sootla, Sten and Gat, Itai and Tan, Xiaoqing Ellen and Adi, Yossi and Liu, Jingyu and Sauvestre, Romain and Remez, Tal and others},
  journal={arXiv preprint arXiv:2308.12950},
  year={2023}
}

@book{kirk2016programming,
  title={Programming massively parallel processors: a hands-on approach},
  author={Kirk, David B and Wen-Mei, W Hwu},
  year={2016},
  publisher={Morgan kaufmann}
}

@article{ke2025qimeng,
  title={QiMeng-MuPa: Mutual-Supervised Learning for Sequential-to-Parallel Code Translation},
  author={Ke, Changxin and Zhang, Rui and Wang, Shuo and Ding, Li and Li, Guangli and Wen, Yuanbo and Zhang, Shuoming and Xu, Ruiyuan and Qin, Jin and Guo, Jiaming and others},
  journal={arXiv preprint arXiv:2506.11153},
  year={2025}
}

@article{liu2026dr,
  title={Dr. Kernel: Reinforcement Learning Done Right for Triton Kernel Generations},
  author={Liu, Wei and Xu, Jiawei and Li, Yingru and Zheng, Longtao and Li, Tianjian and Liu, Qian and He, Junxian},
  journal={arXiv preprint arXiv:2602.05885},
  year={2026}
}

@article{yang2025qwen3,
  title={Qwen3 technical report},
  author={Yang, An and Li, Anfeng and Yang, Baosong and Zhang, Beichen and Hui, Binyuan and Zheng, Bo and Yu, Bowen and Gao, Chang and Huang, Chengen and Lv, Chenxu and others},
  journal={arXiv preprint arXiv:2505.09388},
  year={2025}
}

@inproceedings{hulora,
  title={LoRA: Low-Rank Adaptation of Large Language Models},
  author={Hu, Edward J and Wallis, Phillip and Allen-Zhu, Zeyuan and Li, Yuanzhi and Wang, Shean and Wang, Lu and Chen, Weizhu and others},
  booktitle={International Conference on Learning Representations},
  year={2022}
}

@article{fu2025deep,
  title={Deep think with confidence},
  author={Fu, Yichao and Wang, Xuewei and Tian, Yuandong and Zhao, Jiawei},
  journal={arXiv preprint arXiv:2508.15260},
  year={2025}
}

@article{shao2024deepseekmath,
  title={Deepseekmath: Pushing the limits of mathematical reasoning in open language models},
  author={Shao, Zhihong and Wang, Peiyi and Zhu, Qihao and Xu, Runxin and Song, Junxiao and Bi, Xiao and Zhang, Haowei and Zhang, Mingchuan and Li, YK and Wu, Yang and others},
  journal={arXiv preprint arXiv:2402.03300},
  year={2024}
}

@article{schulman2017proximal,
  title={Proximal policy optimization algorithms},
  author={Schulman, John and Wolski, Filip and Dhariwal, Prafulla and Radford, Alec and Klimov, Oleg},
  journal={arXiv preprint arXiv:1707.06347},
  year={2017}
}

@article{li2025beyond,
  title={Beyond log likelihood: Probability-based objectives for supervised fine-tuning across the model capability continuum},
  author={Li, Gaotang and Qiu, Ruizhong and Chen, Xiusi and Ji, Heng and Tong, Hanghang},
  journal={arXiv preprint arXiv:2510.00526},
  year={2025}
}

@article{wang2026gradients,
  title={Gradients Must Earn Their Influence: Unifying SFT with Generalized Entropic Objectives},
  author={Wang, Zecheng and Liu, Deyuan and Li, Chunshan and Zhang, Yupeng and Zhao, Zhengyun and Chu, Dianhui and Wang, Bingning and Sui, Dianbo},
  journal={arXiv preprint arXiv:2602.11424},
  year={2026}
}

@article{tang2025aligning,
  title={Aligning LLMs with Biomedical Knowledge using Balanced Fine-Tuning},
  author={Tang, Zhenchao and Wang, Fang and He, Haohuai and Zhou, Jiale and Lv, Tianxu and Zhu, Jun and Chen, Shouzhi and Yang, Minghao and Wang, Yu and Wu, Jiayang and others},
  journal={arXiv preprint arXiv:2511.21075},
  year={2025}
}

@article{liu2025deepseek,
  title={Deepseek-v3. 2: Pushing the frontier of open large language models},
  author={Liu, Aixin and Mei, Aoxue and Lin, Bangcai and Xue, Bing and Wang, Bingxuan and Xu, Bingzheng and Wu, Bochao and Zhang, Bowei and Lin, Chaofan and Dong, Chen and others},
  journal={arXiv preprint arXiv:2512.02556},
  year={2025}
}

@article{cao2026qwen3,
  title={Qwen3-coder-next technical report},
  author={Cao, Ruisheng and Chen, Mouxiang and Chen, Jiawei and Cui, Zeyu and Feng, Yunlong and Hui, Binyuan and Jing, Yuheng and Li, Kaixin and Li, Mingze and Lin, Junyang and others},
  journal={arXiv preprint arXiv:2603.00729},
  year={2026}
}

@article{cao2026ascendkernelgen,
  title={AscendKernelGen: A Systematic Study of LLM-Based Kernel Generation for Neural Processing Units},
  author={Cao, Xinzi and Zhai, Jianyang and Li, Pengfei and Hu, Zhiheng and Yan, Cen and Mu, Bingxu and Fang, Guanghuan and She, Bin and Li, Jiayu and Su, Yihan and others},
  journal={arXiv preprint arXiv:2601.07160},
  year={2026}
}

@misc{glm51,
  title        = {GLM-5.1: Towards Long-Horizon Tasks},
  author       = {{Z.ai}},
  year         = {2026},
  month        = {April},
  howpublished = {\url{https://z.ai/blog/glm-5.1}},
  note         = {Official release blog}
}

@misc{nvidiatensorrt,
  author       = {{NVIDIA Corporation}},
  title        = {NVIDIA TensorRT},
  howpublished = {\url{https://developer.nvidia.com/tensorrt/}}
}

@article{massey1951kolmogorov,
  title={The Kolmogorov-Smirnov test for goodness of fit},
  author={Massey Jr, Frank J},
  journal={Journal of the American statistical Association},
  volume={46},
  number={253},
  pages={68--78},
  year={1951},
  publisher={Taylor \& Francis}
}

@article{ye2025flashinfer,
  title={Flashinfer: Efficient and customizable attention engine for llm inference serving},
  author={Ye, Zihao and Chen, Lequn and Lai, Ruihang and Lin, Wuwei and Zhang, Yineng and Wang, Stephanie and Chen, Tianqi and Kasikci, Baris and Grover, Vinod and Krishnamurthy, Arvind and others},
  journal={Proceedings of Machine Learning and Systems},
  volume={7},
  year={2025}
}

@article{miao2025towards,
  title={Towards efficient generative large language model serving: A survey from algorithms to systems},
  author={Miao, Xupeng and Oliaro, Gabriele and Zhang, Zhihao and Cheng, Xinhao and Jin, Hongyi and Chen, Tianqi and Jia, Zhihao},
  journal={ACM Computing Surveys},
  volume={58},
  number={1},
  pages={1--37},
  year={2025},
  publisher={ACM New York, NY}
}


\appendix
\section{Case Study}
We present a case study on the BatchNorm task to further analyze how CuSeT improves functional correctness, as shown in Fig. \ref{fig:case}. The task requires computing column-wise batch normalization over an input tensor \texttt{matA} of shape $(8192, 2048)$, followed by affine transformation using parameters stored in \texttt{matB}.

The SFT-generated program in Fig. \ref{case:sft} is syntactically correct, but it fails to implement the correct execution semantics. Specifically, it assigns each thread to an individual $(n, c)$ element and computes statistics over a spatial dimension, which is inconsistent with the required column-wise reduction. As a result, the incorrect reduction scope leads to degenerate normalization behavior, where the output collapses to near-constant values instead of properly normalized representations.

In contrast, as shown in Fig. \ref{case:cuset}, the CuSeT-generated program maps each thread to a column and performs reduction over the batch dimension, accurately computing mean and variance per column before applying affine transformation. This implementation matches the task semantics, where each column is normalized independently over the batch dimension.

Overall, this case demonstrates that SFT models may produce structurally valid but semantically misaligned kernels due to incorrect execution scope. CuSeT mitigates this issue by better capturing execution-critical CUDA-sensitive patterns, including thread-to-column mapping, reduction scope, and parameter layout, leading to correct functional correctness behavior.

\begin{figure}[!ht]
    \centering
    \begin{subfigure}[t]{\linewidth}
    \begin{tcolorbox}[title=\textbf{BatchNorm}, colback=blue!5, colframe=blue!60!black, width=\linewidth]
    Implement a CUDA kernel for batch normalization. The kernel takes two input tensors: $\texttt{matA}$ of shape $(8192, 2048)$ and $\texttt{matB}$ of shape $(2048, 2048)$, both of float32 type. The output tensor $\texttt{matC\_out}$ has shape $(8192, 2048)$ and float32 type. For each column of $\texttt{matA}$, compute the mean and variance of all elements in that column. Then, for each element in the column, normalize it by subtracting the mean and dividing by the square root of the variance plus $1e^{-5}$. Then, scale the normalized value by a $\gamma$ parameter and add a $\beta$ parameter. The $\gamma$ and $\beta$ parameters for column index $j$ are stored in $\texttt{matB}$: $\gamma$ at the $j$-th element of the first 2,048 elements (flattened) of $\texttt{matB}$, and $\beta$ at the $j$-th element of the next 2,048 elements (flattened) of $\texttt{matB}$. The kernel must be implemented to handle the given tensor dimensions.
    \end{tcolorbox}
        \caption{Task description}
        \label{case:description}
    \end{subfigure}\hfill
    \begin{subfigure}[t]{\linewidth}
        \centering
        \begin{lstlisting}[language=C++, basicstyle=\ttfamily\tiny, breaklines=true, frame=single, escapeinside={(*@}{@*)}]
__global__ void batch_norm_kernel(float* output, const float* input, const float* gamma, const float* beta,
                                 int N, int C, int H, int W, float eps) {
    int idx = blockIdx.x * blockDim.x + threadIdx.x;
    int stride = blockDim.x * gridDim.x;
    
    for (int i = idx; i < N * C; i += stride) {
        int n = i / C;
        int c = i % C;
        
        float mean = 0.0f;
        float var = 0.0f;
        
        for (int hw = 0; hw < H * W; ++hw) {
            int input_idx = n * C * H * W + c * H * W + hw;
            mean += input[input_idx];
        }
        mean /= (H * W);
        
        for (int hw = 0; hw < H * W; ++hw) {
            int input_idx = n * C * H * W + c * H * W + hw;
            float diff = input[input_idx] - mean;
            var += diff * diff;
        }
        var /= (H * W);
        
        float x_hat = (input[n * C * H * W + c * H * W] - mean) / sqrtf(var + eps);
        output[n * C * H * W + c * H * W] = gamma[c] * x_hat + beta[c];
    }
}
        \end{lstlisting}
        \caption{SFT}
        \label{case:sft}
    \end{subfigure}\hfill
    \begin{subfigure}[t]{\linewidth}
        \centering
        \begin{lstlisting}[language=C++, basicstyle=\ttfamily\tiny, breaklines=true, frame=single, escapeinside={(*@}{@*)}]
__global__ void batch_norm_kernel(float* output, const float* input, const float* params,
                                 int N, int C, float eps) {
    int idx = blockIdx.x * blockDim.x + threadIdx.x;
    if (idx >= C) return;

    // Compute mean
    float mean = 0.0f;
    for (int i = 0; i < N; ++i) {
        mean += input[i * C + idx];
    }
    mean /= N;

    // Compute variance
    float var = 0.0f;
    for (int i = 0; i < N; ++i) {
        float diff = input[i * C + idx] - mean;
        var += diff * diff;
    }
    var /= N;

    // Normalize and apply gamma/beta
    float inv_std = 1.0f / sqrtf(var + eps);
    float gamma = params[idx];
    float beta = params[C + idx];
    
    for (int i = 0; i < N; ++i) {
        output[i * C + idx] = gamma * (input[i * C + idx] - mean) * inv_std + beta;
    }
}
        \end{lstlisting}
        \caption{CuSeT}
        \label{case:cuset}
    \end{subfigure}
    \caption{Case study on the BatchNorm task showing that SFT fails to capture correct execution semantics, leading to incorrect normalization scope, while CuSeT produces correct column-wise reduction and functional correct outputs.}
    \label{fig:case}
\end{figure}





\end{document}